# MSMG-Net: Multi-scale Multi-grained Supervised Metworks for Multi-task Image Manipulation Detection and Localization

Fengsheng Wang, Leyi Wei

*Abstract*—With the rapid advances of image editing techniques in recent years, image manipulation detection has attracted considerable attention since the increasing security risks posed by tampered images. To address these challenges, a novel multi-scale multi-grained deep network (MSMG-Net) is proposed to automatically identify manipulated regions. In our MSMG-Net, a parallel multi-scale feature extraction structure is used to extract multi-scale features. Then the multi-grained feature learning is utilized to perceive object-level semantics relation of multi-scale features by introducing the shunted self-attention. To fuse multi-scale multi-grained features, global and local feature fusion block are designed for manipulated region segmentation by a bottom-up approach and multi-level feature aggregation block is designed for edge artifacts detection by a top-down approach. Thus, MSMG-Net can effectively perceive the object-level semantics and encode the edge artifact. Experimental results on five benchmark datasets justify the superior performance of the proposed method, outperforming state-of-the-art manipulation detection and localization methods. Extensive ablation experiments and feature visualization demonstrate the multi-scale multi-grained learning can present effective visual representations of manipulated regions. In addition, MSMG-Net shows better robustness when various post-processing methods further manipulate images.

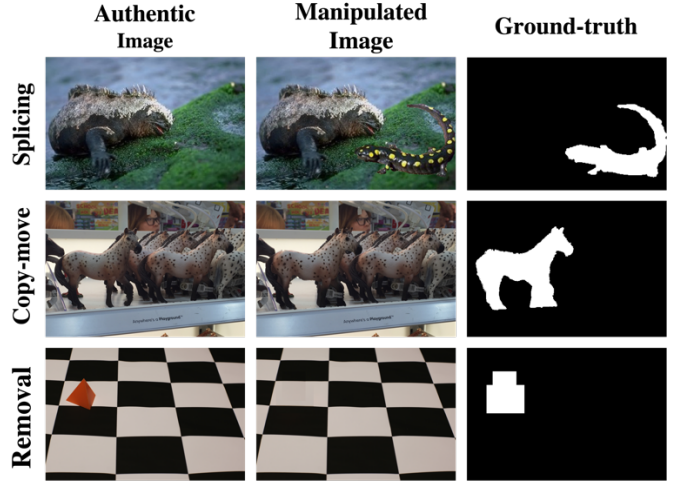

Fig. 1. Example of manipulated images with different manipulation techniques. From the top to bottom are the examples showing manipulations of splicing, copy-move, and removal.

## I. INTRODUCTION

WITH the advances of image manipulation techniques and the merging of user-friendly image editing software, it is easy and effective for common people to produce photo-realistic forgery images at relatively low cost. However, apart from being used in legal applications such as photography, video-game, and virtual reality [1], there are growing concerns on the abuse of the manipulation techniques for various malicious purpose, e.g., fake news, Internet rumors, insurance fraud and blackmail [2]. Therefore, it is crucial to develop effective image manipulation detection methods to examine whether images have been manipulated or not and further locate the tampered regions.

The most widespread and promising image manipulation techniques can be divided into three types: copy-move (copy and move objects from one region to another region in a given image), splicing (copy objects from one image and paste them on another image) and removal (removal of unwanted objects). As shown in Fig. 1, to produce semantically meaningful and perceptually convincing images, these approaches oftentimes introduce or remove special semantic objects in images. These objects often have a complete entity meaning with higher level semantic information. From this perspective, we argue that image manipulation detection and localization just like the game "spot the difference", which means that we must find the subtle differences between two almost same pictures. Usually based on experience, we first find the objects that don't appear together in the both pictures, and then find the other subtle deformed areas by observing the boundary information. Inspired by the game "spot the difference", we consider that one basic task of image manipulation detection and localization is to perceive the different objects in the image and further learning the object-level semantic consistency and inconsistency. Another task is to find the edge artifacts around a tampered region, which contributes to recognize the tampered object and reconstruct

Manuscript submitted November 6, 2022. This work was supported by the Fundamental Research Funds for the Shandong Universities, the National Natural Science Foundation of China [Grant No. 62071278 and No. 62072329]. (Corresponding author: Leyi Wei).

Fengsheng Wang is with School of Software, Shandong University, Shandong, China and Joint SDU-NTU Centre for Artificial Intelligence Research (C-FAIR), Shandong University, Shandong, China (e-mail: wfs@mail.sdu.edu.cn).

Leyi Wei is with School of Software, Shandong University, Shandong, , China and Joint SDU-NTU Centre for Artificial Intelligence Research (C-FAIR), Shandong University, Shandong, China (e-mail: weileyi@sdu.edu.cn).

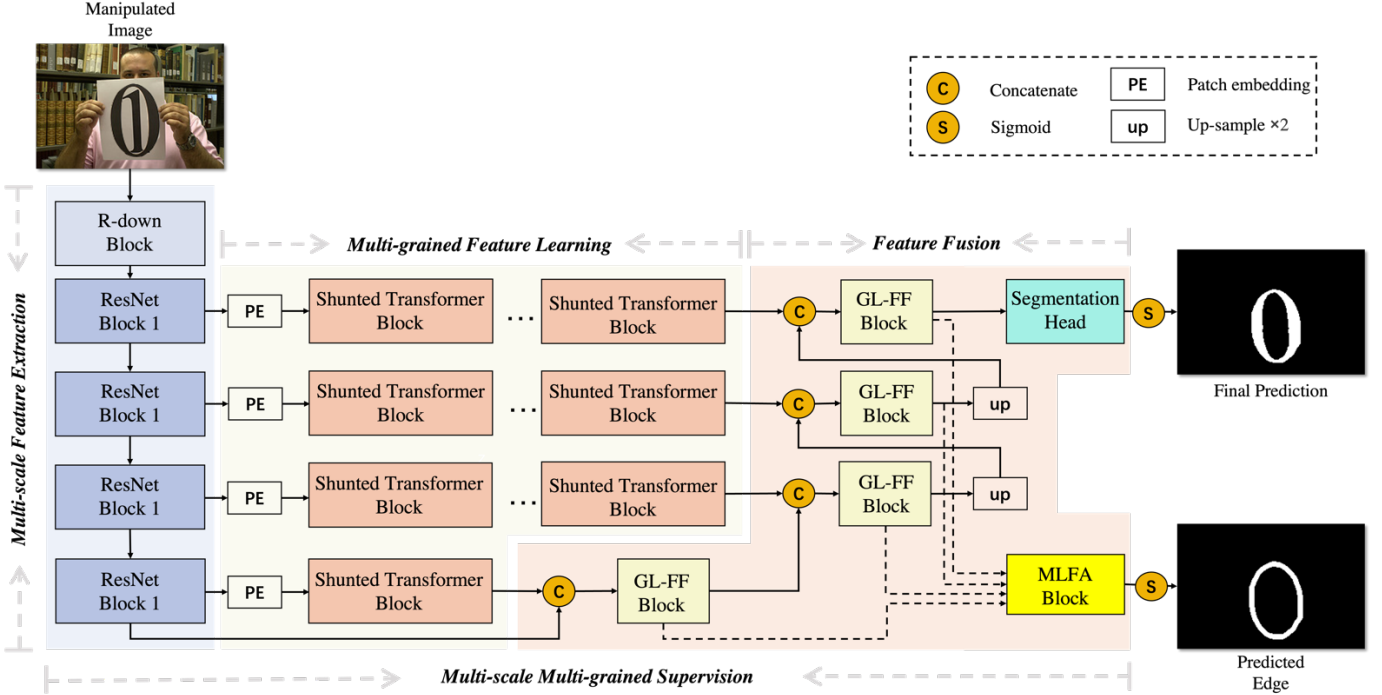

Fig. 2. Conceptual diagram of the proposed MSMG-Net model for image manipulation detection and localization. MSMG-Net consists of multi-scale feature extraction, multi-grained feature learning and feature fusion.

the tampered region. While most of the recent methods mainly use off-the-shelf semantic segmentation network or design ConvNets with the encoder-decoder architecture to recognize manipulated regions [3-7]. Nevertheless, these methods only mitigate the issue of insufficient generalization ability, they are still fundamentally limited by the inadequate feature representations. This in turn naturally raises a question, how to design efficient networks to effectively perceive the objects and encode the edge artifact?

Recent advances in visual recognition are marked by the success of Vision Transformers (ViTs) [8]. Unlike ConvNets, ViTs incorporate the modeling of long-range global contextual information using self-attention to produce image representations. Further, Shunted Transformer [9] model multi-scale objects by learning multi-granularity information. Considering the capability of modeling non-local object relations, attention mechanism is an effective tool to help us capture individual objects, which is very possible as the parts of the manipulation region. However, the attention mechanism could tend to capture those prominent objects and learn their correlations, thereby ignoring the more subtle details in the manipulated regions. For example, the locations of manipulated are the parts of backgrounds such as sky, grassland, and river, which means that the model is more inclined to treat the objects in the foreground as the tampered targets. As manipulating a specific region in a given image inevitably leaves edge artifacts, it is essential to exploit the subtle discrepancy between the tampered regions and its surroundings. ConvNets tend to capture local patterns from tampered regions of edge artifacts, which can be used for solving the edge artifacts detection. However, another issue is that how to combine local processing inherent in convolutions with global processing performed by multi-head self-attention?

In this paper, we design MSMG-Net, an end-to-end multi-scale multi-grained supervised framework for image manipulation detection and localization. The MSMG-Net can be divided into three parts including multi-scale feature extraction, multi-grained feature learning and feature fusion, which focus on capture information from low-level subtle visual artifacts and high-level semantic objects. In multi-scale feature extraction, MSMG-Net first learn coarse-to-fine multi-scale features through resolution down (R-down) block and a stack of ResNet blocks. To exploits the multi-scale potential at a more granular level, we use the shunted transformer block with multi-head shunted self-attention to learn fine-grained features, coarse-grained features and the relation between objects with different sizes of the tampered images in multi-grained feature learning. Finally, given that manipulated region segmentation and edge artifacts detection are intrinsically two distinct tasks, we define a bottom-up Global and local feature fusion (GL-FF) block for manipulated region segmentation and top-down multi-level feature aggregation (MLFA) block for edge artifacts detection in feature fusion. It is noted that the edge artifacts detection is viewed as an auxiliary task to learn fine-grained cues in local regions for improve the results of the main task manipulation segmentation. The framework can be trained in an end-to-end manner by a multi-task supervision strategy that simultaneously process these tasks. To make a fair and head-to-head comparison, we measure the model's performance following the previous studies [10-12]. We conduct experiments on commonly used public image tampering dataset, including CASIA [13], NIST16 [14], Columbia [15], COVERAGE [16] and IMD2020 [17]. The results demonstrate that MSMG-Net outperforms state-of-the-art tampering detection and localization methods. In summary, our major contributions are as follow.

- We propose a novel network MSMG-Net for image manipulation detection and localization. As is shown in Fig. 2, the MSMG-Net contains multi-scale multi-grained supervision architecture designed for learning local visual artifacts and global object-level semantic correlation information in manipulation images and thus obtains more robust and generalizable features.
- We train MSMG-Net with multi-task supervision learning, allowing the model to learn information from semantic contents and edge artifacts in the manipulation images, and consequently improve the model performance substantially.
- Extensive experiments on five benchmark datasets demonstrate that MSMG-Net compares favorably against the state-of-the-art methods.

## II. RELATED WORK

*A. Image Manipulation Detection and Localization.*

Most of related studies focus on a specific type of manipulation, such as copy-move, splicing, and removal. To detect manipulated regions, early works mainly utilize handcrafted or predetermined features such as frequency domain characteristics [18], color filter array (CFA) pattern [19] and the discrete cosine transform (DCT) coefficients [20] to detect manipulated images. However, these methods cannot apply to real forensics because the features are always specific-defined for one type of manipulation techniques. To get more generalized features, deep learning methods are applied to image manipulation detection and localization. ManTra-Net [3] uses predefined filters to capture the manipulation traces and a local anomaly detection network to localize the manipulated regions. CAT-Net [7] constructs two-stream fully convolutional neural network to learn forensic features of compression artifacts on RGB and DCT domains. DenseFCN [4] designs a fully convolutional encoder-decoder architecture by comprising dense connections and dilated convolutions for improving the manipulation localization performance. In order to learn semantic content of the tempered images, SPAN [21] leverage a pyramid architecture and models the dependency of image patches through self-attention mechanisms. SATFL [5] proposes a reliable coarse-to-fine network that utilizes a self-attention mechanism to localize forged regions. In addition, to improve specificity and generalizability, GSR-Net [22] has edge detection as auxiliary task and designs a refinement branch to encourage the learning of boundary artifacts in manipulated regions. Furthermore, MVSS-Net [6] explicitly extract more robust information from boundary artifacts and authentic image by multi-scale supervision. However, these methods only used ConvNets to learn local noise feature without considering object-level semantic information, and thus higher false alarm rate on the manipulated images. In contrast, the proposed method combines local processing inherent in convolutions with global processing performed by multi-head self-attention to fully explore the correlation between tampered regions and non-tampered regions for more accurate image manipulation detection and localization.

*B. Visual Transformer*

Vision Transformer (ViT) [8] is the first work to prove that a pure Transformer can achieve state-of-the-art performance in a variety of computer vision tasks. ViT splits the images into non-overlapped patches (tokens) and perform local self-attention within the token. To model global dependencies via self-attention, Swin Transformer shift the windows over image to get tokens and stack a lot of layers for obtaining a global receptive field. Pyramid Vision Transformer (PVT) designs a spatial-reduction attention to merge tokens of key and query. Furthermore, to capture multi-granularity information and better model objects with different sizes, Shunted-Transformer selectively merges tokens to represent larger object features while keeping certain tokens to preserve fine-grained features. Subsequently, to demonstrate the feasibility of using Transformers in semantic segmentation, SETR [23] designs transformer decoder to extract superior feature representations, achieving impressive performance. After that, Segformer [24] introduces the hierarchical Transformer encoder and lightweight Transformer decoder for further improvement. Inspired by these, we propose a multi-scale multi-grained learning approach, which introduces the visual transformer to learn semantic information in different scales for visual perception and further combines with edge information to capture subtle visual artifacts for image manipulation detection and localization.

## III. METHODOLOGY

*A. Overview of the MSMG-Net*

In this work, we propose a novel model MSMG-Net to predict binary masks of manipulated regions in images. The general framework of our proposed model is illustrated in Fig. 2. Given an RGB image $x$ of size $W \times H \times 3$, the MSMG-Net not only determines whether an image has been manipulated, but also recognizes the edge artifacts around tampered regions by multi-scale multi-grained supervision learning. MSMG-Net firstly down-sample the input images to appropriate resolution size with resolution down (R-down) block. Then the MSMG-Net aggressively learn coarse-to-fine multi-scale features through a stack of ResNet blocks. To exploits the multi-scale potential at a more granular level, we construct tailored Shunted Transformer Block with multi-head shunted self-attention, which effectively capture multi-grained information and implicitly model the object-level semantic representations. By iteratively doing so, MSMG-Net can learn the global multi-scale multi-grained features of non-local correlation of in-the-wild scenarios that involves objects of distinct sizes. Finally, given that manipulated region segmentation and edge artifacts detection two distinct tasks, we employ a multi-task supervision strategy that simultaneously process these tasks. Hence, we define a bottom-up approach for manipulated region segmentation and a top-down approach for edge artifacts detection. The bottom-up approach explicitly fuses multi-scale multi-grained features by the proposed Global and Local Feature Fusion (GL-FF) Block and then implements the segmentation head to classify whether an image and each pixel is tampered or not. For the top-down edge artifacts detection, we introduce the multi-Level feature aggregation (MLA) block to recognize visual artifacts of local regions to make further judgments. We now detail describe the MSMG-Net in the following sections.

### B. Multi-scale Feature Extraction

To provide better feature representations for image manipulation detection and localization, we employ multi-scale feature extraction to learn the feature maps of manipulated images. The multi-scale features provide sufficient semantic and spatial information for subsequent feature fusion. To reduce computational cost, we firstly use resolution down (R-down) block to transform origin image via $3 \times 3$ convolution with stride 2 and max-pool $2 \times 2$ into the feature maps of size $\frac{W}{4} \times \frac{W}{4}$, denoted as $Rdown(x)$.

**ResNet Block:** Several existing image segmentation networks [25, 26] use all high- and low-level features as multi-scale features in the encoder, which include local clues and global contextual information to make the final prediction more reliable. Inspired by this observation, as is shown in Fig. 2, we extract multi-scale feature set $\{f_i, i = 1,2,3,4\}$ from continuous four stacked ResNet blocks of ResNet [27]. The ResNet block structure, illustrated in Fig. 3, consists of the $1 \times 1$ convolution and $3 \times 3$ convolution. The output of multi-scale feature extraction can be expressed by the following equation.

$$\{f_i, i = 1,2,3,4\} \leftarrow ResNet(Rdown(x)) \quad (1)$$

### C. Multi-grained Feature Learning

Consider the visual artifacts and content inconsistencies brought by image manipulation techniques, shunted self-attention [9] is appropriately accounted for learning fine-grained features, coarse-grained features and relation between objects with different sizes of the tampered images. These multi-grained features learn relationships between objects with different sizes in different scales.

It is noted that how to map the feature maps to sequence token affects the performance of manipulation detection. The vanilla ViT [8] split an image into patches and linearly embed each of them. Recent studies [9, 28] demonstrate that using convolution in the patch embedding provides a higher-quality token sequence than a conventional large-stride non-overlapping patch embedding. Therefore, we directly further extract the multi-scale feature $f_i$ through patch embedding (PE) with a $1 \times 1$ convolution and a $3 \times 3$ convolution to map to a reasonable feature channel dimension. Then we feed these features into shunted transformer block to learn multi-scale multi-grained features.

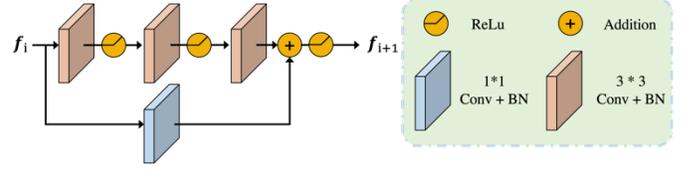

Fig. 3. Illustration of the ResNet block.

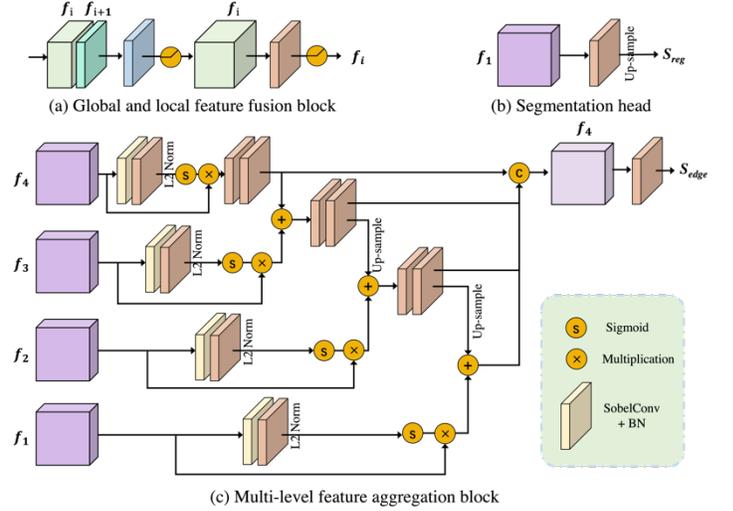

Fig. 4. Elements in the feature fusion module. (a) Global and local feature fusion (GL-FF) block fuses multi-scale feature maps. (b) Segmentation head is utilized to obtain the final segmentation detection results. (c) Multi-level feature aggregation (MLFA) block is utilized to generate the edge map.

**Shunted Transformer Block:** The shunted transformer block is made up of shunted self-attention and multi-layer perception (MLA) block. It is noted that LayerNorm (LN) is applied before every block, and residual connections after every block. The shunted self-attention is designed to capture multi-grained objects especially the small-size ones [9]. To extract feature of local clues and global contextual information of tampered images, we use shunted self-attention to learn semantic relationship at different scales. We firstly flatten the multi-scale features to a sequence of 2D vector with additional positional embeddings as the input embedding. Then the shunted transformer is utilized to capture different granularity information of the multi-scale features. The complete process can be formulated as:

$$f_i = Flatten(f_i), i = 1, 2, 3, 4 \quad (2)$$

$$f_i = LN(ShuntedTransformer(f_i)) + f_i \quad (3)$$

$$f_i = LN(MLA(f_i)) + f_i \quad (4)$$

TABLE I

TRAINING AND TESTING IMAGES OF SIX DATASETS USED IN OUR EXPERIMENTS. THE SYMBOL – INDICATES INFORMATION UNAVAILABLE AND THE SYMBOL \ INDICATES MANIPULATION TECHNIQUES UNUSED.

| Dataset | Training-testing split | | The categories of manipulation techniques | | | Total |
|---|---|---|---|---|---|---|
| | Training | Testing | copy-move | splicing | removal | |
| CASIA | 5123 | 920 | 3733 | 2310 | \ | 6043 |
| COVER | 75 | 25 | 100 | \ | \ | 100 |
| Columbia | \ | 180 | \ | 180 | \ | 180 |
| NIST16 | 404 | 160 | 68 | 288 | 208 | 564 |
| IMD2020 | 1610 | 400 | - | - | - | 2010 |

Finally, we restore the embedding to resolution size of origin feature maps, which then serves as the input of feature fusion in our MSMG-Net.

### D. Feature Fusion

To better output final unified feature representations of manipulated region and the corresponding edge artifacts, we design global and local feature fusion block and multi-level feature aggregation block to fuse different scale features.

**Global and Local Feature Fusion (GL-FF) Block:** Inspired by the U-Net architecture [29], we use a naive method to integrate global and local features from multi-grained feature learning. As is shown in Fig. 4(a), the features $f_i$ and $f_{i+1}$ are directly concatenated and further fused by the $1 \times 1$ convolution and $3 \times 3$ convolution. We progressively fuse these features by bottom-up approach. Then we feed the feature $f_1$ into segmentation head to generate the response map $S_{seg}$ as the final prediction in Fig. 2.

**Multi-Level Feature Aggregation (MLFA) Block:** Several works have shown that edge artifacts information can provide useful constraints to guide feature extraction for image manipulation detection and localization [6, 22]. Thus, considering multi-scale multi-grained features preserve multi-level sufficient edge information, we introduce a top-down multi-level feature aggregation block (MLFA) to explicitly learn the edge artifact representation, illustrated in Fig. 4(c). The feature set $\{f_i, i = 1,2,3,4\}$ go through the Sobel convolution for identifying edge regions [6]. To enhance the interactions across these features, we aggregate them via element-wise addition. And an additional $3 \times 3$ convolution is applied after the element-wise addition. After repeating this operation three times, we obtain the fused feature via channel-wise concatenation. Then we bilinearly up-sample $4 \times$ the results to generate the edge response map $S_{edge}$ as the predicted edge in Fig.2.

### E. Multi-Task Supervision Strategy

We define two losses correspond to special multi-task targets, i.e., a manipulated region segmentation loss for improving the model's sensitivity and an edge loss for learning subtle edge artifacts for improving the model's robustness of manipulation detection and localization.

**Manipulated region segmentation loss.** As manipulated region typically in minority in a given image, the manipulation detection and localization can be viewed as extreme data imbalanced problem. Therefore, we use the Dice loss as manipulated segmentation loss, which can be denoted as $loss_{reg}$.

$$loss_{seg} = 1 - \frac{2\sum_{i,j} S_{seg_{i,j}} \cdot y_{i,j}}{\sum_{i,j} S_{seg_{i,j}} + \sum_{i,j} y_{i,j}} \quad (4)$$

where $y_{i,j} \in (0,1)$ denotes the binary ground-truth indicating whether the pixel $(i,j)$ is manipulated or not.

**Edge artifacts detection loss.** For edge artifacts around tampered regions, we again use the Dice loss for detection, denoted as $loss_{edge}$. Since edge artifacts detection is an auxiliary task, following previous studies [6], we directly compute the loss edge obtained by the top-down approach, which retains the resolution size of the original image with $4 \times$ down-sampling, which reduces computational cost during training and improves the detection performance.

**Combined loss.** We use a convex combination of the two losses as the total loss of the propose MSMG-Net, which is formulated as follows.

$$loss_{seg} Loss = \gamma_r \cdot loss_{seg} + \gamma_e \cdot loss_{edge} \quad (4)$$

where $\gamma_r$ and $\gamma_e$ are weights of manipulated region segmentation loss and edge artifacts detection loss.

## IV. EXPERIMENTS

### A. Experimental Settings

**Datasets.** Following previous studies [6, 10-12], we compare and evaluate our method with current state-of-the-art methods on Nist Nimble 2016 [14] (NIST16), CASIA [13], COVERAGE [16], Columbia [15] and IMD2020 [17] dataset. These datasets are widely used in the image manipulation detection and localization.

- NIST16 [14] is a challenging dataset which contains all three tampering techniques. The manipulations in this dataset are post-processed to conceal visible traces. They also provide ground-truth tampering mask for evaluation.

TABLE II
QUANTITATIVE PERFORMANCE COMPARISON RESULTS OF MSMG-NET WITH FIVE STATE-OF-THE-ART METHODS ON CASIA, NIST, COLUMBIA, COVER, IMD2020.

| Method | CASIA | | NIST16 | | Columbia | | COVER | | IMD2020 | |
|---|---|---|---|---|---|---|---|---|---|---|
| | AUC | F1 | AUC | F1 | AUC | F1 | AUC | F1 | AUC | F1 |
| Mantra-Net | 0.648 | 0.223 | 0.667 | 0.207 | 0.618 | 0.263 | 0.611 | 0.210 | 0.785 | 0.265 |
| CAT-Net | 0.704 | 0.203 | 0.751 | 0.173 | 0.693 | 0.277 | 0.753 | 0.288 | 0.786 | 0.239 |
| DenseFCN | 0.631 | 0.209 | 0.762 | 0.261 | 0.617 | 0.336 | 0.728 | 0.334 | 0.723 | 0.286 |
| SATFL | 0.697 | 0.246 | 0.829 | 0.287 | 0.658 | 0.399 | 0.767 | 0.347 | 0.796 | 0.300 |
| MVSS-Net | 0.748 | 0.390 | 0.821 | 0.441 | 0.795 | 0.524 | 0.811 | 0.418 | 0.817 | 0.411 |
| MSMG-Net | 0.726 | 0.425 | 0.831 | 0.492 | 0.808 | 0.567 | 0.853 | 0.480 | 0.877 | 0.494 |

TABLE III
PERFORMANCE COMPARISON OF MSMG-NET SETUPS IN THE ABLATION STUDY. EACH SETUP IS EVALUATED ON NIST16 DATASET.

| Ablation part | Setup | Component | | | | | AUC | F1 |
|---|---|---|---|---|---|---|---|---|
| | | Scale #1 | Scale #2 | Scale #3 | Scale #4 | Loss | | |
| Multi-scale Feature Extraction | Setup #0 | ✗ | ✗ | ✗ | ✗ | ✗ | 0.778 | 0.355 |
| Multi-grained Feature Learning | Setup #1 | ✓ | ✗ | ✗ | ✗ | ✗ | 0.697 | 0.205 |
| | Setup #2 | ✗ | ✓ | ✗ | ✗ | ✗ | 0.811 | 0.386 |
| | Setup #3 | ✗ | ✗ | ✓ | ✗ | ✗ | 0.801 | 0.372 |
| | Setup #4 | ✗ | ✗ | ✗ | ✓ | ✗ | 0.846 | 0.392 |
| | Setup #5 | ✓ | ✓ | ✓ | ✓ | ✗ | 0.801 | 0.478 |
| - | Full setup | ✓ | ✓ | ✓ | ✓ | ✓ | 0.831 | 0.492 |

- CASIA [13] provides spliced and copy-moved images of various objects. The tampered regions are carefully selected and some post-processing techniques like filtering and blurring are also applied. CASIA is composed of CASIAv1 and CASIAv2 splits. Samples from both subsets are provided binary ground-truth masks.
- Coverage [16] dataset contains relatively small dataset generated by copy-move techniques, and the ground-truth masks are also available.
- Columbia [15] dataset focuses on splicing based on uncompressed images. Ground-truth masks are provided.
- IMD2020 [17] includes real-life manipulated images as well as manually created ground truth masks, which contains the most diverse quantization tables.

For a fair and direct comparison with the current state-of-the-art methods, we re-trained and evaluated these models on CASIA, NIST16, COVER and Columbia datasets following the most popular training-testing splitting configurations in [10-12]. For IMD2020, we choose an approximate 4:1 training-testing ratio. The Details of these datasets are presented in Table 1.

**Evaluation Criteria.** Following previous works [3-6], we compute pixel-level F1-score (F1) and the Area Under the receiver operating characteristic Curve (AUC) as our evaluation metrics for performance comparison of manipulation detection and localization.

Note that previous works commonly report performance with the decision threshold selected per test set, allowing one to compare models under their optimal conditions. However, this setting leads to overly optimistic performance estimates, as in practice, a model's decision threshold (or its operating point) has to be pre-specified and fixed. Because in data without GT, the most appropriate threshold cannot be predicted, we use the median value of 0.5 as the threshold to determine the positive and negative classes for overall methods.

**Baseline Models.** We evaluate and compare MSMG-Net with five currently published SoTA methods i.e., ManTra-Net[1] [3], CAT-Net[2] [7], DenseFCN[3] [4], SATFL[4] [5] and MVSS[5] [6]. These compared methods have publicly available source codes. For a fair and reproducible comparison, we have re-trained the above models and evaluated them using the related parameters optimally assigned or automatically chosen as described in the reference papers.

---

[1] https://github.com/RonyAbecidan/ManTraNet-pytorch
[2] https://github.com/mjkwon2021/CAT-Net
[3] https://github.com/ZhuangPeiyu/ Dense-FCN-for-tampering-localization
[4] https://github.com/tansq/SATFL
[5] https://github.com/dong03/MVSS-Net

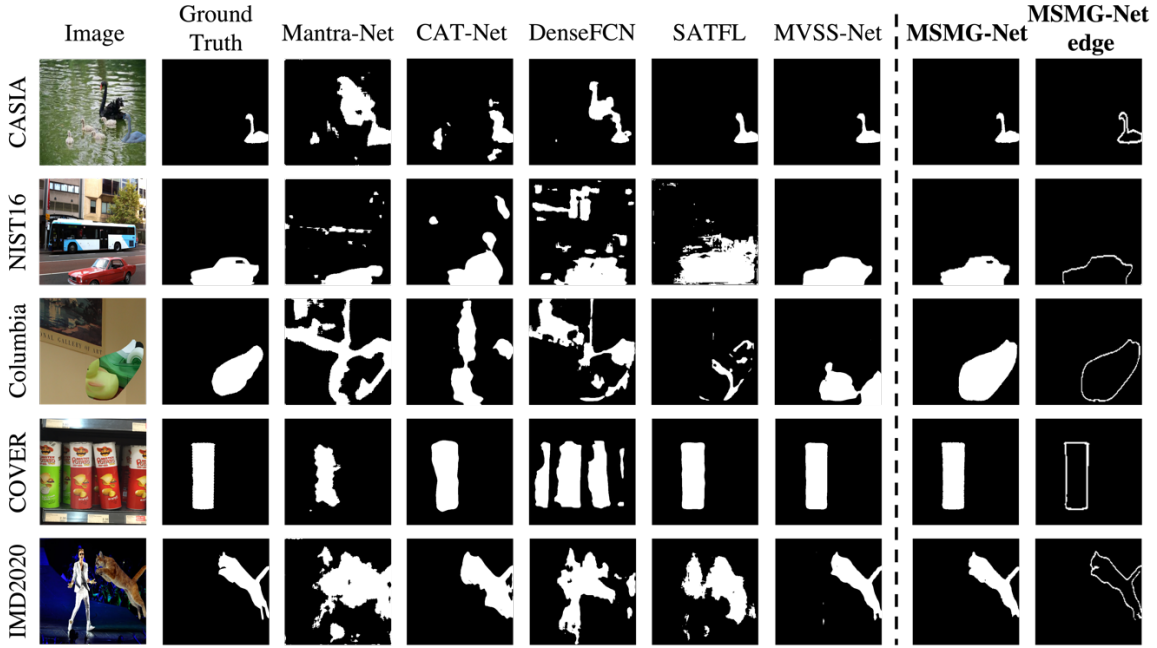

Fig. 5. Visual qualitative comparison results by MSMG-Net with five state-of-the-art methods on five datasets. From left to right, we show manipulated images, ground-truth binary masks of manipulated images, the manipulation detection and localization results of the Mantra-Net, CAT-Net, DenseFCN, SATFL, MVSS-Net, and MSMG-Net, and the predicted edge artifacts of MSMG-Net.

**Implementation.** The proposed network is implemented in PyTorch and trained on an NVIDIA Tesla V100 GPU. The input size is 512×512. The Resnet Block used in multi-scale feature extraction are initialized with ImageNet-pretrained counterparts. We use an Adam optimizer with a learning rate periodically decays from $10^{-4}$ to $10^{-6}$. We apply regular data augmentation for training, including flipping, blurring, and compression.

### B. Segmentation Comparison Results with the State-of-the-arts

*1) Quantitative Results:* To compare the manipulation detection and localization, we consider five state-of-the-art methods Mantra-Net, CAT-Net, DenseFCN, SATFL and MVSS-Net. Quantitative results are shown in Table 2. As can be seen, the proposed MSMG-Net outperforms Mantra-Net, CAT-Net, DenseFCN, SATFL and MVSS-Net in terms of AUC and F1 by a large margin. We attribute this improvement to the ability of capturing information from low-level subtle visual artifacts and high-level semantic objects of our multi-scale multi-grained supervision learning. More specifically, the spatial relation in images, ignored in SoTA methods, guarantee the generalization ability of MSMG-Net. Moreover, the constraint of subtle edge artifacts explicitly enhances the boundary information and ensures the fineness of predicted segmentation results, whereas other methods fail to fully exploit details of edge information around tampered regions.

*2) Quantitative Results:* The manipulation detection and localization results, shown in Fig. 5, indicate that our MSMG-Net outperforms the baseline methods remarkably. Specifically, MSMG-Net yields segmentation results that are close to the ground truth with much less mis-segmented tissue. In contrast, Mantra-Net, CAT-Net, and DenseFCN give unsatisfactory results, where a large number of mis-segmented regions exist. SATFL improves the results, but the performance is still not promising. The boundary detection results of ManTra-Net, Cat-Net, DenseFCN and SATFL are far less precise than that of MSMG-Net, because they neglect edge artifact information. In Contrast, MVSS-Net and MSMG-Net employ edge supervision branches to learn edge features for enhancing the performance of manipulation and localization. However, as can be observed in column 8 of Fig. 5 (especially the image in the second row), the MSMG-Net get more precise prediction results. The accurate segmentation of MSMG-Net is owed to our multi-scale multi-grained supervision strategy, which better learns semantic consistency and inconsistency in images. This strategy introduces shunted self-attention mechanism to perceive objects and learn their relationship in images. By perceiving the objects of different sizes and recognizing the edge artifacts, the MSMG-Net finally obtain satisfactory results.

### C. Ablation Study

To reveal the influence of the multi-scale multi-grained supervision, we conduct ablation study to evaluate the MSMG-Net in varied setups with the components added progressively. The quantitative results are shown in Table 3 and the qualitative visualization are observed in Fig. 6. Corresponding to setup #1, setup #2, setup #3 and setup #4, the scale #1, scale #2, scale #3 and scale #4 represent the features of multi-grained learning according to the scale from shallow to deep. And setup #5 fuses the multi-scale multi-grained features. In addition, the loss as an alternative strategy is used for learning edge artifacts. The All results reported in this section are obtained with NIST16 dataset.

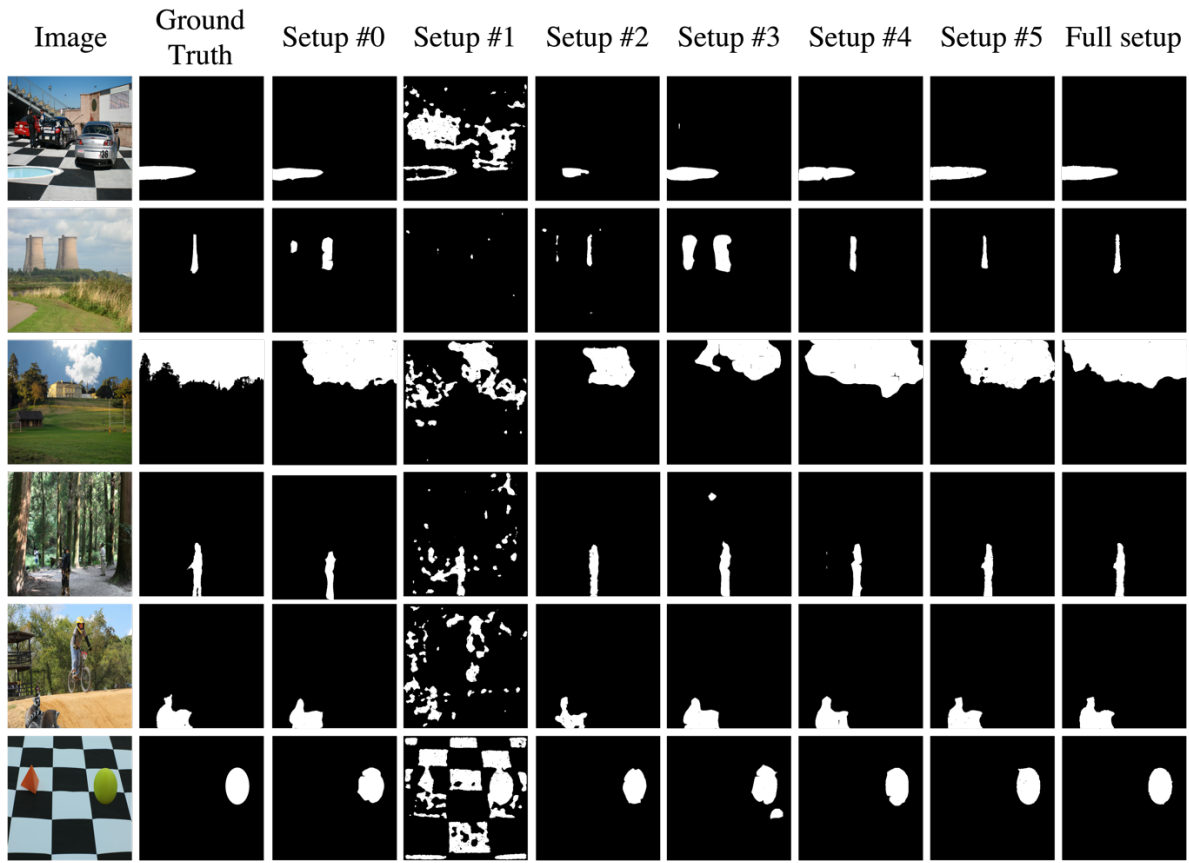

Fig. 6. The visualization results of MSMG-Net in the ablation study on the multi-scale multi-grained learning and edge artifacts detection. Among all the setups, full setup achieves more fine segmentation results.

**Influence of multi-scale feature extraction.** The setup #0 directly adopts the final Resnet block output of multi-scale feature extraction to predict the final manipulation region. The overall performance of setup #0 is lower than seg-all in Table 3, and the results clearly demonstrate the superiority of the multi-scale multi-grained features. The seg-all, comparing with setup #1, setup #2, setup #3 and setup #4 in Table 3 and Fig. 6, perform the best, showing the significance of the individual component in different scales.

**Influence of multi-grained feature learning.** To further demonstrate the necessity of the multi-grained feature learning, we make a differences comparison of every individual component in different scales in Fig. 6. Especially in the face of complex backgrounds in the third rows, the results differ greatly. It is worth mentioning that the seg #1 achieves poor performance in terms of AUC and F1. The possible reason is that the scale #1 focus on local edge information and introduce more noise pattern, resulting in the weak feature representation learning ability. This is a side-by-side verification that multi-grained features can focus on different aspects of an image.

**Influence of edge artifacts detection loss.** Comparing Seg-all and full setup, we can see a clear slight improvement in AUC and F1, suggesting that adding edge artifacts detection loss makes the model effective for manipulation detection. This change is not only confirmed by higher AUC and F1, but is also confirmed in the ninth column of Fig. 6. In contrast, full setup gives relatively clear boundaries, especially in the subtle infection regions. From the left to right, the results demonstrate how the propose model in varied setups obtains more accurate results.

In conclusion, the great performance of MSMG-Net is owed to multi-scale multi-grained supervision strategy, where the parallel shunted transformer blocks are utilized to learn multi-scale multi-grained features to locate manipulated regions, and then the edge artifacts are employed for further fine segmentation, which learn sufficient information of manipulated images.

*D. Feature Visualization*

In order to further present the information of multi-scale multi-grained features, we provide comprehensive visualization results of feature maps to analyze the differences in visual representations in Fig. 7. These feature maps contain the shunted self-attention feature maps of final four shunted transformer block of the parallel structure in the multi-grained learning and the response maps of the segmentation head in feature fusion.

We observe that the feature maps capture the desired information including local structure (edges, lines, textures, etc) and global abstract semantics progressively from the scale #1 to the scale #4. We clearly see that the low-level spatial feature maps extracted from scale #1 and scale #2 highlight both general region boundary and local details, which gradually replaced by global object-level semantic features shown in scale #3 and scale #4. It demonstrates that the MSMG-Net not only learns global dependencies, but also pays more attention to important local patterns in multi-scale multi-grained learning. With the in-depth view of network, MSMG-Net identifies approximate tampered areas and further introduce the GL-FF block and MLA block to enhance the edge representations learned from manipulated images with involvement of more local edge artifacts details. We can visually feel the obvious difference between the tampered and non-tampered areas, which reflects the fact that our model learns object-level semantic consistency and inconsistency. As a result, the final response maps yield fine-grained prediction results, which demonstrates that it is more effective to extract feature maps from the parallel structure of multi-scale multi-grained learning in our MSMG-Net.

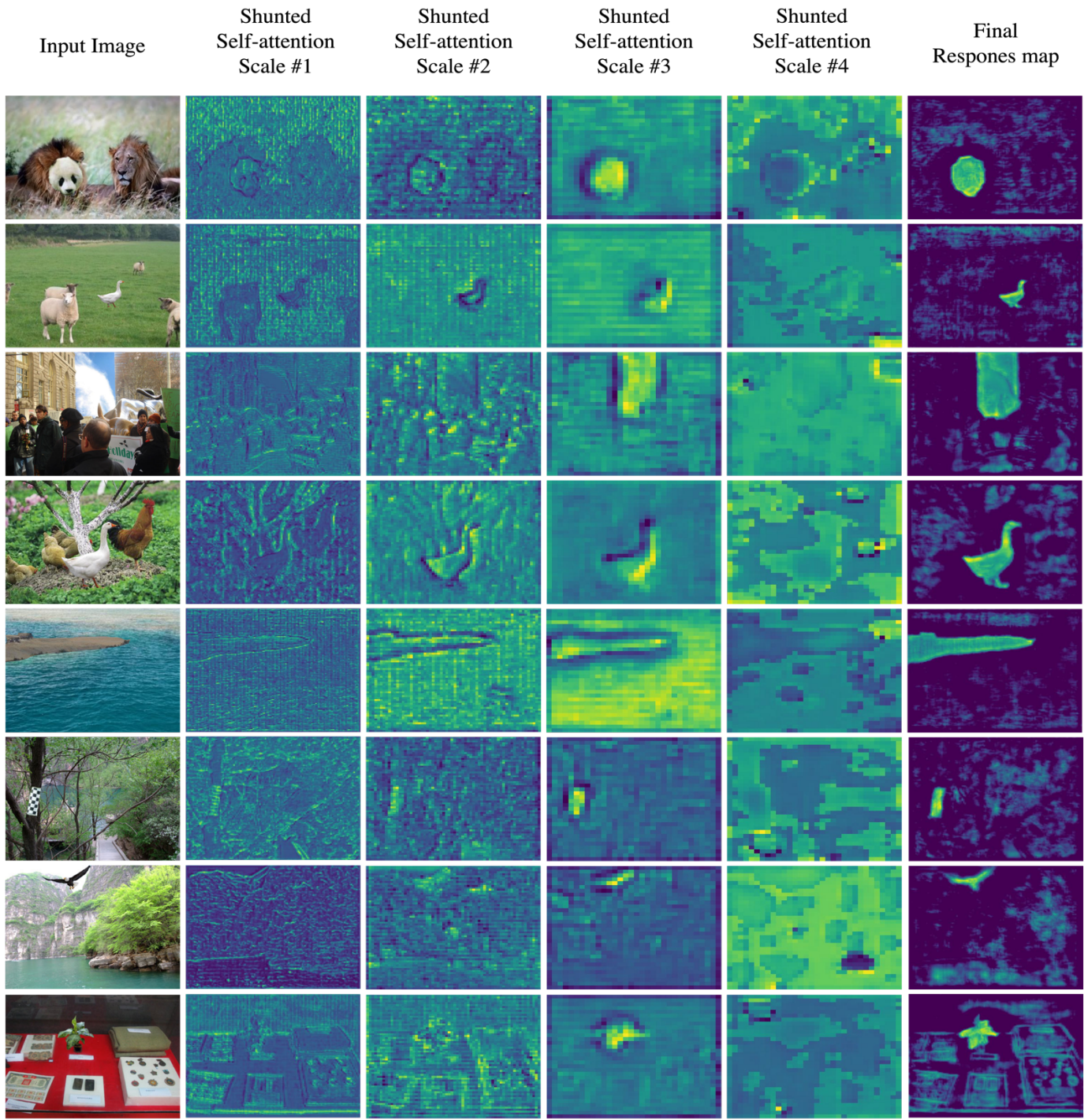

Fig. 7. The visualization results of the shunted self-attention feature maps of shunted transformer block in the multi-grained learning and the response maps of the segmentation head in feature fusion.

*E. Robustness Analysis*

Following previous studies [3, 6, 11], we evaluate the model robustness against four image post-processing methods, Gaussian blur, Gaussian noise, JPEG compression and ISO noise over NIST dataset to verify the robustness of MSMG-Net. The detailed results of robustness analysis are shown in Fig. 8. For each post-processing method, we vary the kernel size in Gaussian blur (from 3 to 9), variance of Gaussian noise (from 3 to 9), quality in JPEG compression (from 50% to 100%), and variance of ISO noise (from 0.05 to 0.2) for comprehensive evaluation. As can be observed, Gaussian blur affects the detection performance more severely, in particular when a larger kernel size of 9×9, which blurs images and erases manipulation traces around tampered regions. In addition, compared with other baselines, MSMG-Net achieves the most general robust performance on ISO noise. The results of MSMG-Net is owed to our multi-scale multi-grained learning, where a parallel partial shunted transformer block designed to learn coarse-to-fine manipulation segmentation features. In summary, our model, MSMG-Net, consistently performs the best among all methods and can effectively tackle the challenges brought by various post-processing methods.

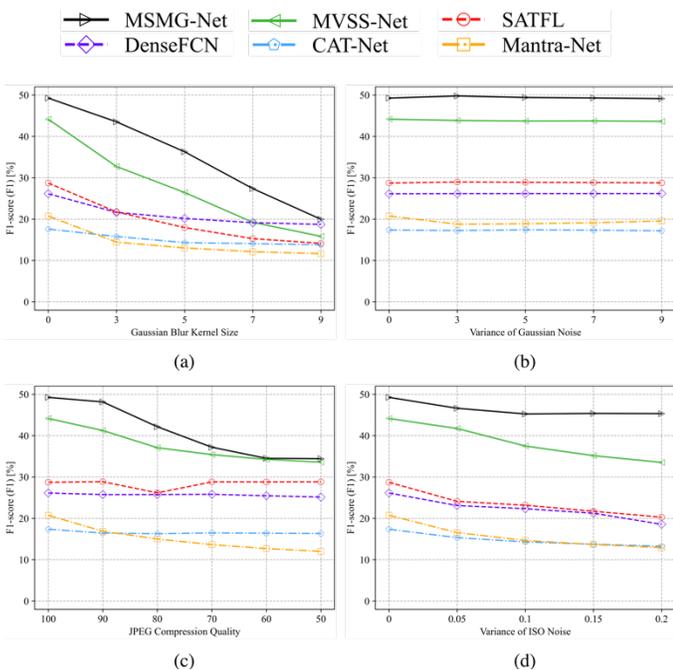

Fig. 8. Robustness evaluation against four image post-processing techniques, i.e. Gaussian blur, Gaussian noise, JPEG compression and ISO noise over NIST dataset.

*E. Limitation analysis*

Given the challenging nature of the task, failure is inevitable when MSMG-Net is applied to real scenarios. As is shown in Fig. 9(a), facing three types of manipulated techniques including removal, splicing and copy-move, MSMG-Net gives some unsatisfactory results. When the artifacts in tampered regions are the subtleties with additional post-processing methods [30], MSMG-Net obtains incomplete tampered areas owing to insufficient tampering traces, which has been a difficult point for various manipulated detection methods [1]. In addition, Fig. 9(b) shows the prediction results for authentic images in CASIA dataset. The propose MSMG-Net identifies areas that are clearly different from surrounding environment. Especially for the third row, the artistic effect Bokeh is used in the upper part of the image to highlight the main object of interest on the photo by blurring all out-of-focus areas [31], leading to the false alarms. In summary, the proposed method MSMG-Net shows good performance on extracting precise object boundaries and visually salient edges, and perceiving the semantic consistency and inconsistency between different objects and backgrounds in the pictures. Moreover, the MSMG-Net accurately identifies the complete objects in tampered images and can be used as an auxiliary tool for image manipulation detection and localization.

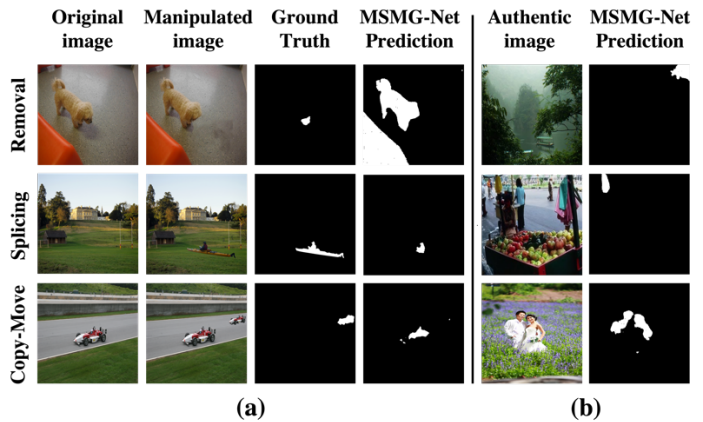

Fig. 9. The failure cases of the proposed MSMG-Net. (a) Manipulated images in NIST16 dataset. (b) Authentic images in CASIA dataset.

## V. CONCLUSION

In this paper, we have proposed a novel image manipulation detection and localization deep network, named MSMG-Net, which utilizes multi-scale multi-grained supervision to improve the identification of tampered regions. MSMG-Net designs a parallel structure to lean multi-scale multi-grained features, and fuse these features by a bottom-up approach for manipulated region segmentation and a top-down approach for edge artifacts detection. Extensive experiments on five benchmark datasets demonstrate the effectiveness of proposed method for image manipulation detection and localization, outperforming the current methods. Extensive ablation experiments and feature visualization demonstrate the superiority of the multi-scale multi-grained features. In addition, the proposed method also exhibits better robustness against Gaussian blur, Gaussian noise, JPEG compression and ISO noise. With the great performance of MSMG-Net, we believe it can be an effective tool in image manipulation detection and localization.